\documentclass[journal]{vgtc}                    

\onlineid{0}

\vgtccategory{Analytics \& Decisions}

\title{GFlowState: Visualizing the Training of Generative Flow Networks Beyond the Reward}

\author{%
  Florian Holeczek,
  Andreas Hinterreiter,
  Alex Hernandez-Garcia, 
  Marc Streit, and
  Christina Humer
}

\authorfooter{
  \item
  	Florian Holeczek, Andreas Hinterreiter, and Marc Streit are with Johannes Kepler University Linz.
    E-mail: florian.holeczek@posteo.at, marc.streit@jku.at, andreas.hinterreiter@jku.at.
  \item
  	Christina Humer is with Eidgenössische Technische Hochschule Zürich.
  	E-mail: christina.humer@inf.ethz.ch.

  \item Alex Hernandez-Garcia is with Mila – Quebec AI Institute.
  	E-mail: alex.hernandez-garcia@mila.quebec.
}

\abstract{%
We present \emph{GFlowState}, a visual analytics system designed to illuminate the training process of Generative Flow Networks (GFlowNets or GFNs).
GFlowNets are a probabilistic framework for generating samples proportionally to a reward function. While GFlowNets have proved to be powerful tools in applications such as molecule and material discovery, their training dynamics remain difficult to interpret.
Standard machine learning tools allow metric tracking but do not reveal how models explore the sample space, construct sample trajectories, or shift sampling probabilities during training.
Our solution,\emph{GFlowState}, allows users to analyze sampling trajectories, compare the sample space relative to reference datasets, and analyze the training dynamics.
To this end, we introduce multiple views, including a chart of candidate rankings, a state projection, a node-link diagram of the trajectory network, and a transition heatmap.
These visualizations enable GFlowNet developers and users to investigate sampling behavior and policy evolution, and to identify underexplored regions and sources of training failure.
Case studies demonstrate how the system supports debugging and assessing the quality of GFlowNets across application domains.
By making the structural dynamics of GFlowNets observable, our work enhances their interpretability and can accelerate GFlowNet development in practice.
}


\teaser{
  \centering
  \includegraphics[width=\linewidth, alt={A view of a city with buildings peeking out of the clouds.}]{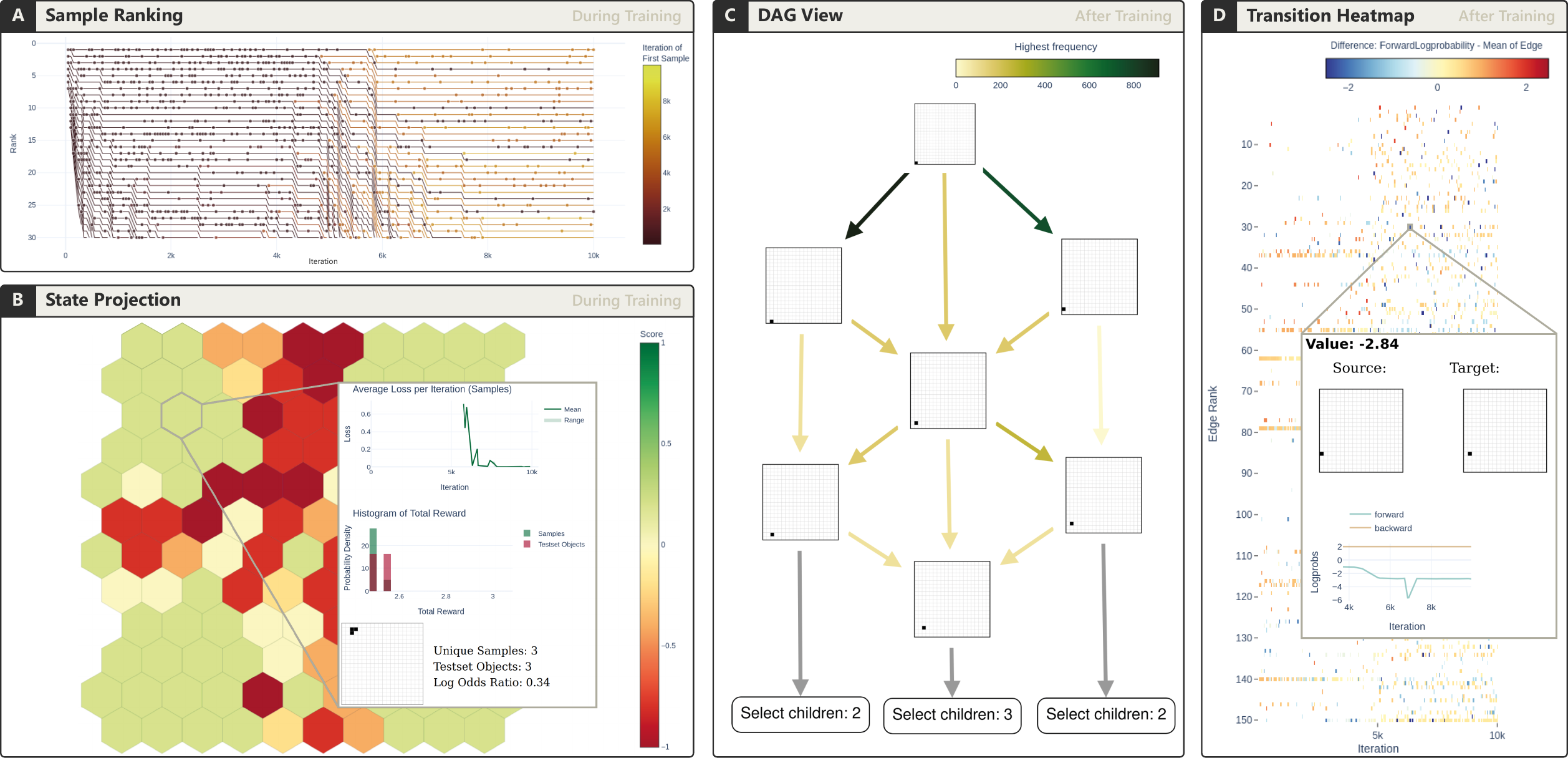}
  \caption{\emph{GFlowState} allows users to analyze the training behavior of GFlowNets based on four views: (A)~A ranking of generated objects based on their reward. (B)~A~visualization of the sample space of generated objects relative to a reference dataset. (C)~A directed acyclic graph representing the generated trajectories. (D)~A transition heatmap indicating transition probabilities. While components A and B depend on the samples produced during training, components C and D analyze the graph formed by their trajectories after training has completed.
  }
  \label{fig:teaser}
}

\graphicspath{{figs/}{figures/}{pictures/}{images/}{./}} 

\usepackage{mathptmx}                  
\usepackage{amsfonts}                  
\usepackage{enumitem}
\usepackage{calc}

\newcommand{\task}[1]{\textsf{\small\bfseries#1}}

\begin{document}


\firstsection{Introduction}

\maketitle

Scientific discovery increasingly relies on generative AI models to suggest novel candidates, such as molecules or materials~\cite{wang2023scientific, jain2023gflownets}. 
Accelerating this discovery process is critical to addressing global challenges such as climate change and global health by advancing battery technologies, electrocatalysts for hydrogen generation~\cite{rolnick2022tackling}, and the design of therapeutic molecules such as antibodies for combating infectious diseases~\cite{dara2022machine}.
Large-scale initiatives in machine learning for scientific discovery have demonstrated the potential of data-driven approaches to explore vast chemical and material spaces~\cite{chanussot_open_2021}.
However, these spaces are combinatorially large, making an exhaustive search infeasible, and motivating algorithms that can efficiently explore and prioritize promising candidates.

\emph{Generative Flow Networks} (GFlowNets or GFNs for short)~\cite{bengio2021flow} were designed to tackle this task by learning to sample objects proportionally to a reward function.
This principle, combined with the generalization potential of deep learning, enables an effective exploration of the search space and the generation of many diverse high-value solutions. 
GFlowNets have shown promising results in electrocatalyst~\cite{podina_catalyst_2025} and antibody design~\cite{yin_synergy_2025}, offering advantages over other generative algorithms.
For example, compared to Markov Chain Monte Carlo (MCMC) methods, GFlowNets can learn the structure of the sample space rather than relying solely on local transitions, amortizing the sampling cost during training.
Compared to standard reinforcement learning, GFlowNets are particularly well-suited for tasks that require generating many diverse, high-reward solutions instead of a single optimal one.

However, GFlowNet training remains difficult to interpret and diagnose.
Key characteristics of GFlowNet training, such as sample space coverage, state transition dynamics, and the relationship between sampling probability and reward, are difficult to analyze with existing visualization platforms.
The underlying structures rapidly grow combinatorially large, exceeding the limits of manual inspection and conventional monitoring tools.
This creates a critical need for tailored visual analytics methods that make the structure and dynamics of GFlowNet training observable.
In this work, we introduce \emph{GFlowState}, a tool designed for interactive exploration and diagnosis of the GFlowNet training processes. 

To the best of our knowledge, no prior work has addressed the specific analytical needs for understanding and diagnosing GFlowNet training dynamics.
Although recent advances in GFlowNet research have focused on improving learning algorithms and applications~\cite{shen2023understanding, mistal_crystal-gfn_2023, jain2023gflownets}, tools for interactive analysis of their learning behavior remain largely unexplored.
As a result, practitioners currently rely on aggregate metrics or ad-hoc inspection methods that provide limited insight.

In contrast, the visualization community has demonstrated the effectiveness of model-specific visual analytics solutions for understanding complex machine learning models.
In reinforcement learning, interactive visualization systems enable the analysis of trajectories, policy decisions, and the comparison between agents during training~\cite{metz_visitor_2023, zahavy_graying_2016}.
Similar tailored tools have also been proposed for other model classes, including sequence and language models~\cite{strobelt_lstmvis_2018, strobelt_seq2seq-vis_2019, sevastjanova_visual_2023}, as well as convolutional neural networks~\cite{wang_cnn_2021}.
These approaches demonstrate that visualization tools designed to capture the structure and training dynamics of a specific model class can reveal behaviors that are difficult to observe with standard metrics alone.
However, these approaches are tightly coupled to the structure of the respective learning paradigms and cannot be directly applied to GFlowNets.

To address this gap, we collaborated with GFlowNet developers at Mila, the Quebec AI Institute, to identify analysis tasks to rigorously assess the quality of GFlowNet training runs, inspect model shortcomings, and aid debugging.
Key tasks include (\textit{i})~the analysis of sampling trajectories, (\textit{ii})~ the exploration of the space of generated training samples, and (\textit{iii})~the analysis of training dynamics throughout episodes.
Based on an iterative feedback loop with our collaborators, we designed \emph{GFlowState} to address these and other tasks effectively.

Our main contributions are:
\begin{itemize}
    \item an identification of analysis tasks for improving the understanding and diagnosis of GFlowNet training;
    \item \emph{GFlowState}, an interactive prototype addressing these tasks; and
    \item an evaluation of the approach based on case studies from two application domains.
\end{itemize}

\section{A Generative Flow Networks Primer}
\label{sec:background}

GFlowNets provide a generative framework for sampling proportionally to a reward function.
GFlowNets treat sampling as a sequential decision problem: Generating the final sample is a process of constructing it from smaller building blocks.
To illustrate these concepts, we first introduce the grid environment used in the original GFlowNet paper~\cite{bengio2021flow}.
Throughout the paper, we will use this grid environment as a guiding example. 

\begin{figure}[t]
  \centering
  \includegraphics[width=0.7\columnwidth, alt={A heatmap of a 20 by 20 grid showing the higher reward near the corners. Arrows indicate a trajectory starting at (0,0) and leading to (14,4).}]{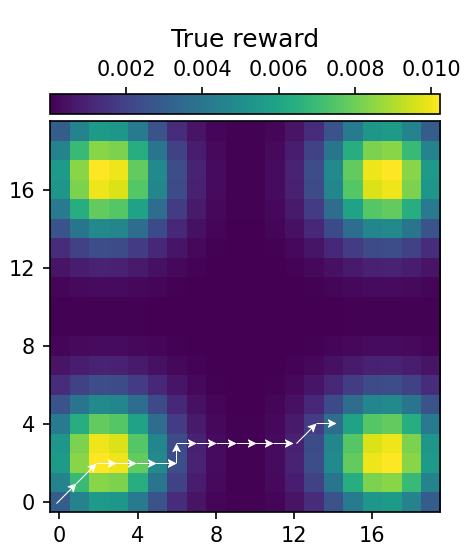}
  \caption{A heatmap of the true rewards for the grid environment used as an example, along with an exemplary trajectory. High-reward modes are placed near the corners of the grid. The white trajectory is one possible way to sample the state~(14,4).
  }
  \label{fig:reward}
\end{figure}

Consider a Markov decision process (MDP) where the states are cells of a two-dimensional grid of side length $H$.
The agent starts at position $s_0 =(0,0)$ and moves by sampling actions $a = (a_x,a_y); a_x, a_y \in [0,1]$.
The current state transitions into the next state according to the sampled action: $s_{i+1} = s_i + a_i$. The agent can also sample a \textit{stop} action to terminate the sampling trajectory.
In this example setup, a fixed reward distribution is defined to place high-reward modes near the corners of the grid (see~\cref{fig:reward}).

The final object is the cell on the grid at which the agent samples the \textit{stop} action; all previous actions that together lead to this point define the trajectory.
The state space $s \in \mathcal{S}$ of all trajectories forms a directed acyclic graph (DAG) with $(0,0)$ as the root node or source state, common to all of them.
The terminal states $x \in \mathcal{X}$ (i.e., the final objects) are leaf nodes in the DAG.
In general, not all intermediate states are part of the sampling space; in other words, the \textit{stop} action may not be valid in all states.
Querying the reward function with the final generated object $x$ yields object-reward pairs $(x, R(x))$ that can be used to train the so-called policy model.

Given a specific state, the action to get to the next state is sampled using a \emph{forward policy} model.
The policy is typically implemented as a neural network that takes a state as input and outputs \emph{flow} values for the possible next actions.
From these flow values, a distribution can be constructed that allows sampling the next states.
By interpreting the DAG as a flow network in which the flow is a nonnegative scalar at each node and edge, the model can be trained with a suitable objective.
In the original GFlowNet paper, Bengio et al. (2021)~\cite{bengio2021flow} proved that if the flow entering a state is the same as the flow leaving it, for all states in $\mathcal{S}$ (known as \emph{Flow Consistency}), the forward policy will sample objects proportionally to their reward.
This is the main theorem of GFlowNets and explains the diversity of the samples.
Other learning strategies (e.g., trajectory balance~\cite{malkin2022trajectory}) require a \emph{backward} policy in addition to the forward policy. While the forward policy gives a distribution over all \emph{children} of a state, the backward policy gives a distribution over all \emph{parents} of a state. 
To learn more about the theory behind GFlowNets, see Bengio et al. (2023)~\cite{bengio2023gflownet} or, for an interactive introduction, the \emph{GFlowNet Playground}~\cite{holeczek_gflownet_2025}.

In summary, GFlowNets construct sequences of states and actions, modeling action probabilities as flows that adhere to specific constraints.
This way, GFlowNets can sample final states proportionally to their reward function.
As a consequence, GFlowNets can generate sets of high-reward candidates that better capture the diversity of the solution space.

\section{Related Work}

We situate our work within three related areas: visual analytics systems for machine learning, experiment tracking tools, and graph exploration methods.

\subsection{Visual Analytics for Machine Learning}

Visualization and visual analytics tools are routinely used to analyze machine learning models both during and after the training process~\cite{hohman_visual_2019,yuan2021survey}.
Based on the interrogative survey by Hohman et al~\cite{hohman_visual_2019} survey, we situate our work as a visual analytics tool for \emph{debugging and improving models}~(WHY) aimed at \emph{model developers and builders}~(WHO).

Similarly situated works with a focus on training dynamics have been developed for the purpose of understanding, diagnosing, and steering models during training~\cite{yuan2021survey}.
For example, InstanceFlow~\cite{puhringer_instanceflow_2020} and ConfusionFlow~\cite{hinterreiter_confusionflow_2022} monitor the evolution of model class confusion at a local and global level, respectively.
While being agnostic to the model type, they are tailored towards classification tasks.
For generative tasks, \emph{GAN Lab}~\cite{kahng_gan_2019}, is an educational tool to learn about and experiment with generative adversarial models.
For diffusion models, \emph{EvolvED}~\cite{prasad_evolved_2025} introduces evolutionary embeddings to explain the generation process.

Reinforcement learning is an ML paradigm more closely related to GFlowNets.
Although the learning tasks differ---GFlowNets are generative frameworks, while reinforcement learning involves navigating an agent in an environment---the overall conceptualization of the two approaches is similar: both paradigms involve an environment in which solutions are formed through states, actions, and rewards.
Many of the visualization solutions targeting reinforcement learning~\cite{zahavy_graying_2016,wang_visual_2022} use two-dimensional projection methods to explore the states that the agents encounter over the episodes.
VISITOR~\cite{metz_visitor_2023} augments a similar state space projection by showing transitions between states using lines.
The PolicyExplainer~\cite{mishra_why_2022}, on the other hand, proposes a less technical interface targeted at non-expert users. 

Although we can build on parts of these works, the analytics needs of GFlowNet developers differ and are primarily based on the generated samples (see Section~\ref{sec:tasks}).
Specifically for GFlowNets, we found little literature focused on improving understanding.
Shen et al.~\cite{shen2023understanding} studied how to investigate and fix flow issues, but their framework remains purely computational, with no user-facing visualizations. Holeczek et al.~\cite{holeczek_gflownet_2025}, on the other hand, approach GFlowNet understanding from a completely different perspective, providing an interactive educational article that aims to give readers a high-level intuition about GFlowNet components, training, and applications.

\subsection{Experiment Tracking Tools} 

The standard for assessing the quality of GFlowNet training runs, as inferred from interviews with GFlowNet developers, is to use experiment-tracking platforms such as TensorBoard~\cite{tensorflow_developers_tensorflow_2026} or Weights \& Biases~\cite{biewald_experiment_2020}.
These tools allow users to log and view metrics, such as losses and rewards, monitor changes in the weights and biases of the neural nets used for the policy, and create simple custom plots.
These are essential aspects to track during training, enabling deeper analysis beyond aggregated high-level metrics.
However, these tools do not capture the specific architectural characteristics of GFlowNets.
GFlowNets typically rely on two neural networks, one for their forward and one for their backward policy.
Although existing tools allow inspection of these networks, assessing the progress of GFlowNet training requires additional aspects: (\textit{i})~the correlation of sampling probabilities and rewards, (\textit{ii})~the diversity and novelty of samples, and (\textit{iii})~the coverage of the sample space by the samples relative to a dataset.
We designed \emph{GFlowState} to address these gaps, helping developers estimate how well their models performs on these aspects.
This deeper exploration of the learning process allows users to identify where and why models may struggle.

\subsection{Graph Exploration}
\label{sec:graph_exploration}

A~specific task that \emph{GFlowState} aims to address is the exploration of the vast DAG of the training trajectories.
One way to achieve this is to start with the source node and let the user interactively expand the DAG, a strategy referred to as bottom-up filtering~\cite{von_landesberger_visual_2011}.
TreePlus by Lee et al.~\cite{lee_treeplus_2006} is an early example of enabling users to expand a tree by interactively selecting nodes of interest.
Juniper~\cite{nobre2019juniper} extended this concept to multivariate graphs, where the nodes carry rich attribute data. Rather than showing the full graph, users query for nodes or subgraphs of interest, which are then visualized as a linearly laid out spanning tree. This linear layout enables integration with an attached table and an adjacency matrix, supporting tasks such as attribute comparison across nodes while preserving the graph's topological structure.

These approaches have also proven to be effective for exploring language model output.
The beam search tree~\cite{spinner_-generaitor_2024, spinner_revealing_2025} shows subgraphs of the tree spanned by the possible outputs of a language model.
It can support the analysis, explanation, and adaptation of generated text.
As tree expansion has been shown to work for large combinatorial spaces, such as language model output, we adapt it to the DAG structure of GFlowNet training trajectories, thereby making the trajectory space accessible and analyzable.

\section{User Tasks}
\label{sec:tasks}

GFlowNets were first proposed by researchers at Mila -- the Quebec AI Institute\footnote{\url{https://mila.quebec/en}}, which remains a leading center for research on the framework.
Through expert interviews with researchers actively developing GFlowNets at Mila, we identified key needs to improve the assessment and debugging of GFlowNet training.
We found three groups of tasks in GFlowNet development: analysis of sample trajectories~(\task{Traj}), analysis of generated samples~(\task{Samp}), and analysis of the training dynamics~(\task{Dyn}).

\subsection{Analyzing Sampling Trajectories}
\label{sec:task:sampling_trajectories}

Sampling trajectories are defined by the sequence of actions (or transitions) the model takes to produce a single sample.
The visualization of the GFlowNet MDP and the DAG, or parts of them, constructed from multiple sampling trajectories, can provide insights into the model's decision process. For example, it can reveal transitions that occur frequently under specific circumstances or not at all.
This gives rise to two tasks in this group:
\begin{enumerate}[
    label=\task{Traj~\arabic*},
    labelindent=0pt,
    labelwidth=\widthof{\task{Traj~2}},
    leftmargin=!,
    ]
    \item \label{task:traj-steps} Tracking the model's sampling behavior and decisions at each step of a trajectory; and
    \item \label{task:traj-factors} Identifying factors that lead to the generation of a sample.
\end{enumerate}

\subsection{Analyzing generated samples}
\label{sec:task:sample_space}

GFlowNet developers want to understand the learned sampling policy of the model. However, in most relevant problems, the sample space is combinatorially large, and studying the sampling probabilities on the entire space is intractable.
Instead, they can analyze a sample of generated candidates in a variety of ways.
For debugging purposes, they are interested in finding regions where the model fails to sample proportionally to the reward.
This can be done by examining the relationship between forward sampling probabilities and the corresponding reward values.
Deviations in proportionality indicate areas where the learned policy is not yet aligned with the target distribution.
After identifying such a region of interest in the sample space, analyzing individual samples can help understand model weaknesses or training issues.
To explore the range of promising candidates, developers want to identify high-reward regions that are modeled accurately, as indicated by low training loss.
Together, this gives rise to three tasks:
\begin{enumerate}[
    label=\task{Samp~\arabic*},
    labelindent=0pt,
    labelwidth=\widthof{\task{Samp~3}},
    leftmargin=!,
    ]
    \item \label{task:space-prop} Validate that the sampling is performed proportionally to the reward function;
    \item \label{task:space-region} Locate regions of high and low reward or loss; and
    \item \label{task:space-outlier} Identify samples that are outliers in one of the points above (i.e., no proportional sampling, or outlying reward or loss) for further analysis.
\end{enumerate}

Given the intractability of the full sample space, one way to assess whether a trained GFlowNet approximately samples proportionally to the reward is by estimating the likelihood of sampling the objects in a reference dataset.
We refer to this dataset as the \emph{validation set}.
The validation set consists of objects representative of the GFlowNet sample space, with rewards computed using the same reward function used in training.
The probability of sampling an object in a dataset can be estimated through importance sampling, by generating multiple backward trajectories from the object.
The sampling probabilities can then be compared to the rewards. For example, a typical summary metric is the correlation between sampling (log) probabilities and (log) rewards, where higher correlation indicates that the GFlowNet may sample proportionally to the reward.
The validation set should ideally cover the full range of the reward function (not just high-reward objects) that the model is expected to model.
In the grid environment example, the validation set contains all grid points.
This gives rise to a fourth task in this task group:
\begin{enumerate}[
    label=\task{Samp~\arabic*},
    resume,
    labelindent=0pt,
    labelwidth=\widthof{\task{Samp~3}},
    leftmargin=!,
    ]
    \item \label{task:space-valid} Compare the sample distribution with a reference distribution from a validation set.
\end{enumerate}
The validation set, if available, can also be used to inform the outlier analysis as defined in~\ref{task:space-outlier}.

\subsection{Analyzing Training Dynamics}
\label{sec:task:training_dynamics}

As the policy model learns, the flow between states shifts, and consequently, the transition probabilities change.
Developers need to identify and analyze these shifts to understand how sampling behavior changes over time.
This gives rise to the first task in this group:
\begin{enumerate}[
    label=\task{Dyn~\arabic*},
    labelindent=0pt,
    labelwidth=\widthof{\task{Dyn~3}},
    leftmargin=!,
    ]
    \item \label{task:dynamic-shifts} Analyze shifts in transition probabilities during training.
\end{enumerate}

Second, developers are interested in understanding which regions of the sample space are discovered during training and how exploration dynamics evolve over time:
\begin{enumerate}[
    label=\task{Dyn~\arabic*},
    resume,
    labelindent=0pt,
    labelwidth=\widthof{\task{Dyn~3}},
    leftmargin=!,
    ]
    \item \label{task:dynamic-expl} Analyze exploration dynamics in the sample space over time.
\end{enumerate}
This analysis can inform adjustments to hyperparameters that promote more effective exploration.

Finally, when a model discovers a mode that yields a higher reward, it may exclusively sample from it, thereby stopping exploration.
This is referred to as \emph{mode collapse}.
In the grid environment, an example of mode collapse is when the model discovers the high-reward mode in the lower-left corner, while disregarding the other three modes.
It is vital for model developers to identify iterations in which mode collapse occurs:
\begin{enumerate}[
    label=\task{Dyn~\arabic*},
    resume,
    labelindent=0pt,
    labelwidth=\widthof{\task{Dyn~3}},
    leftmargin=!,
    ]
    \item \label{task:dynamic-collapse} Detect mode collapse.
\end{enumerate}
Successfully performing this task can help developers fix the underlying problem to ensure that the GFlowNet creates a more diverse sample of candidates.

\section{GFlowState Design}

Based on the tasks identified in Section~\ref{sec:tasks}, we introduce \emph{GFlowState}, a visual analytics solution analyzing the training process of GFlowNets.
\emph{GFlowState} features four main views (see \cref{fig:teaser}): the \emph{Sample Ranking}, the \emph{State Projection}, the \emph{DAG View}, and the \emph{Transition Heatmap}.
In the following sections, we describe the design of these views, their interplay, and the analytical tasks they address.

\subsection{Workflow and Interaction}
\label{sec:interaction}

\begin{figure}[t]
  \centering
  \includegraphics[width=0.95\columnwidth, alt={Workflow of GFlowState.}]{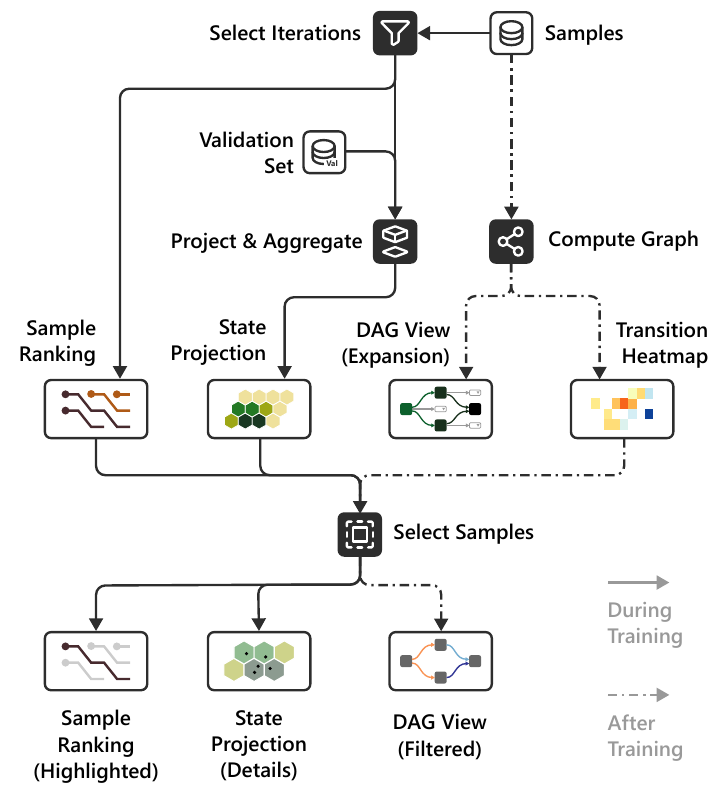}
  \caption{Workflow of \emph{GFlowState}. Arrows show the component dependencies during and after training. All visualizations use the training samples as their data source, filtered by the user-selected iteration range. The \emph{DAG View} and the \emph{Transition Heatmap} use the graph computed from the trajectories of the samples. The \emph{State Projection} also uses a validation set, if available, and projects and aggregates the data into hexbins. Selecting samples or trajectories also highlights them or shows their details in other visualizations.
  }
  \label{fig:workflow}
\end{figure}

Figure~\ref{fig:workflow} illustrates the workflow of \emph{GFlowState}. During model training, samples and their attributes are written to a database. 
Optionally, a validation set can serve as a reference for the generated training samples.
To facilitate a focused analysis of the exploration dynamics across iterations (\ref{task:dynamic-expl}), a central slider allows users to select the range of iterations; all subsequent visualizations use only samples within this selected range.

The \emph{Sample Ranking} shows the highest- and lowest-ranked samples in a bump chart based on a chosen metric.
To construct the \emph{State Projection} view, dimension reduction is applied to the samples and the validation set (if provided) based on their features.
The resulting two-dimensional projection is aggregated in hexagonal bins.
While the \emph{Sample Ranking} and the \emph{State Projection} can be used during GFlowNet training, other visualizations require computing the DAG derived from the samples.
Although this is possible during training, it is computationally expensive and therefore disabled by default.

After training is complete, the DAG is calculated over all sample trajectories.
The \emph{Transition Heatmap} shows the highest-ranking transitions across different metrics and the iterations in which they occur.
In the \emph{DAG View}, Users can explore the DAG of the sample trajectories by expanding it interactively from the root node.
To allow a detailed inspection of samples of interest (\ref{task:space-outlier}), users can select samples or transitions of interest for further analysis.
Selections in any of the views update the other views as indicated in Figure~\ref{fig:workflow}.

In all charts, hovering over samples or transitions visualizes their associated state, enabling quick identification.
The state representation can be customized for the specific use case (see \cref{sec:impl} for more information on environment integration).
In the \emph{DAG View}, the state is displayed directly in the node.
Further metrics or attributes may also be displayed on demand (see \cref{sec:dag} for more details).

\subsection{Sample Ranking}
\label{sec:ranking}

\begin{figure*}[t]
  \centering
  \includegraphics[width=0.9\linewidth, alt={Sample ranking showing the cumulative highest reward samples over all iterations.}]{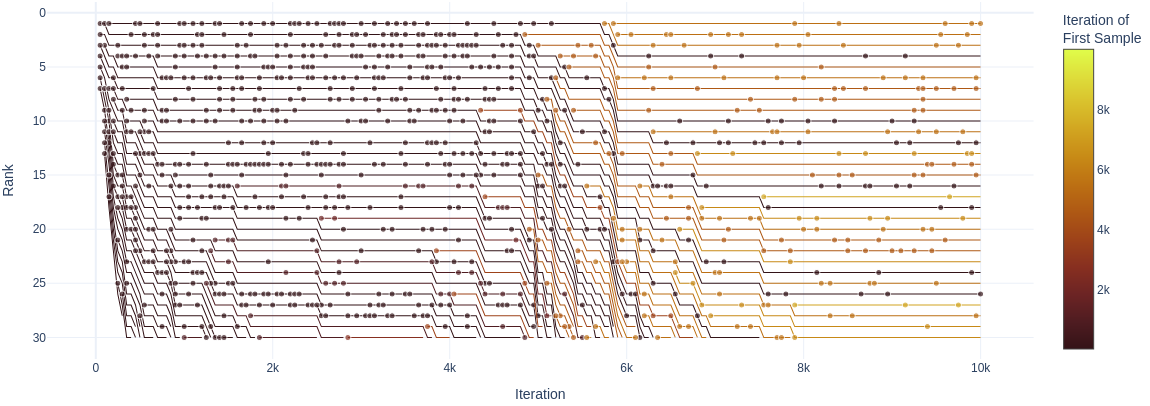}
  \caption{\emph{Sample Ranking} showing the cumulative highest reward samples over all iterations. Note the change around iteration 4,500, when the model discovers two new high-reward modes that displace the previously high-ranked samples.
  }
  \label{fig:ranking}
\end{figure*}

The \emph{Sample Ranking} view serves as a first overview of the training dynamics (see (\cref{fig:ranking}).
This view displays the top $N$ final objects, ranked by a metric as a bump chart across the iterations.
Users can select to rank by reward, loss, or a custom metric
The \emph{Sample Ranking} allows users to select samples of interest for further exploration in other visualizations and to identify iterations in which the model discovers new, highly ranked samples.

For each iteration, the highest- or lowest-ranked objects of this and previous iterations are displayed.
A~line shows the change in rank over time for each unique sample.
Vertical changes in the lines indicate a successful exploration of the model towards new high-reward areas (\ref{task:dynamic-expl}), as many new promising candidates join the ranking.
An example of this can be seen in Figure~\ref{fig:ranking} around iteration 4,500.
Regions with mostly flat, horizontal lines point towards little exploration---a potential indicator of mode collapse (\ref{task:dynamic-collapse}).
Selecting samples in the \emph{Sample Ranking} highlights them in the \emph{State Projection} and shows their trajectories in the \emph{DAG View}.

\subsection{State Projection}
\label{sec:projection}

\begin{figure}[t]
  \centering
  \includegraphics[width=\columnwidth, alt={The state projection of the grid environment as a hexagonal grid. Coloring indicates a higher reward in the grid's corners.}]{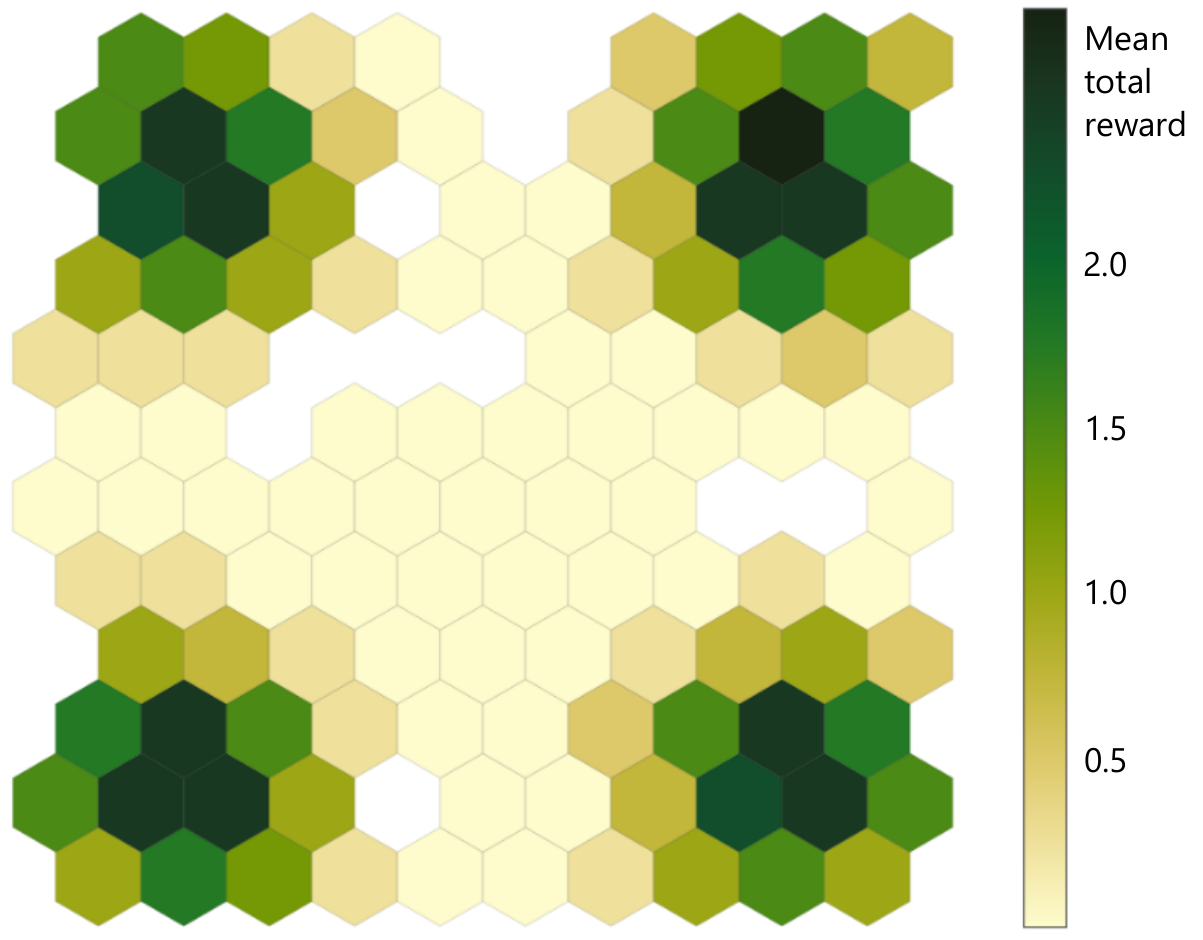}
  \caption{\emph{State Projection} of the grid environment. Points representing explored states are aggregated in a hexagonal grid. Since the environment is two-dimensional, no dimension reduction was applied in this example. For more advanced environments, a two-dimensional UMAP projection is performed using custom feature representations. Here, the color encodes the average reward per hexagonal bin. Other metrics can be selected to explore different properties of the state space. 
  }
  \label{fig:state_space}
\end{figure}

The \emph{State Projection} view allows users to get a sense of the explored sample space and the reward landscape (see \cref{fig:state_space}).
To construct the \emph{State Projection}, dimension reduction is applied to all sampled objects, either based on their states or using customizable feature extraction.
If available, objects from a validation dataset can be projected together with the training samples and displayed as a reference.
The projected points are binned on a hexagonal grid, and aggregation metrics are computed per bin.
This allows comparison of areas in the projection space while avoiding visual clutter and overplotting, thereby reducing cognitive load.
In addition to the typical training metrics, \emph{GFlowState} offers metrics based on reward--probability correlation and validation set coverage.
We outline the various metric choices below.

\textbf{Correlation of Reward and Sampling Probability.} This metric shows the correlation between the probability of sampling an object and its reward.
Sampling proportionally to the reward is a defining property of a correctly trained GFlowNet, enabling it to sample diverse yet high-value candidates.
Coloring bins by this metric helps users to identify regions where the model deviates from this objective (\ref{task:space-prop}). 
Hovering over a bin also displays the non-aggregated scatterplot of sampling probability versus reward, enabling detailed inspection regardless of the sample count.
The correlation between sampling probability and reward is an important metric in GFlowNet training that is usually analyzed globally.
By showing this metric on a local level, researchers can assess model quality in more detail (\ref{task:space-prop}).

\textbf{Training Metrics.} Bins can be colored by reward, loss, or other custom metrics, allowing identification of regions with especially high or low values (\ref{task:space-region}).
For example, regions with high loss may be of particular interest, as in this case the model fails to sample proportionally to the reward.
This indicates that the model does not correctly represent this part of the sample space and needs additional training to do so.
Similarly, high-reward regions indicate where to sample high-value candidates.

\textbf{Validation Set Coverage.} If a validation dataset is provided, bins can also be colored according to the ratio between the number of sampled objects and the validation set objects.
To account for differences in total sample size, the odds ratio (i.e., a measure of how strongly the two sets are associated) is computed and scaled to the range $[-1, 1]$.
Here, a value of 0 indicates that the ratio of the number of generated training samples and the number of objects in the validation set in a bin is the equal to the global ratio between generated samples and items in the validation set.
A~value of $+1$ indicates that a bin contains only validation set objects, while $-1$ indicates only generated training samples.
Bins with low values may be particularly informative, as they indicate regions represented in the validation set but underrepresented in the model output.
This allows for assessing the quality of the model by validating that it is able to correctly sample from the distribution of the validation set (\ref{task:space-valid}).

Regardless of the metric chosen for the color encoding, users can select a bin to inspect its individual samples.
Selecting one of the bins focuses on its location and reveals all individual samples as points.
The selected samples are also highlighted in the \emph{Sample Ranking}, and their trajectories are shown in the \emph{DAG View}.
Alternatively, the whole \emph{State Projection} can be displayed as a scatterplot, providing information about the iteration at which the sample was generated and its loss or reward value.
Selected samples are also highlighted in the \emph{Sample Ranking} and their trajectories are displayed in the \emph{DAG View} (\ref{task:space-outlier}).

Hovering over a bin shows its attributes in detail.
Users can inspect the loss over all iterations (\ref{task:dynamic-expl}), the distribution of rewards, and the number of samples and validation set entries.
Furthermore, a visual summary of the states of the samples is shown.
This summary of states highly depends on the environment.
For the grid environment example, the summary is simply a visual representation of all samples within the grid (see the bin information in \cref{fig:teaser}B).
This summary visualization is customizable.
In molecule generation or other graph-based environments, for example, the maximum common substructure might be a good summary, indicating which structure is shared across samples within a bin.
In continuous environments, density plots can be used (similar to the projection summaries proposed by Eckelt et al.~\cite{eckelt_visual_2022}).
In Section~\ref{sec:impl} we provide details on how the environment-specific requirements are integrated into our system.

\subsection{DAG View}
\label{sec:dag}

\begin{figure}[h]
  \centering
  \includegraphics[width=\columnwidth, alt={A Directed Acyclic Graph with visualizations of the grid environment states as nodes. Node handles allow selecting a node children.}]{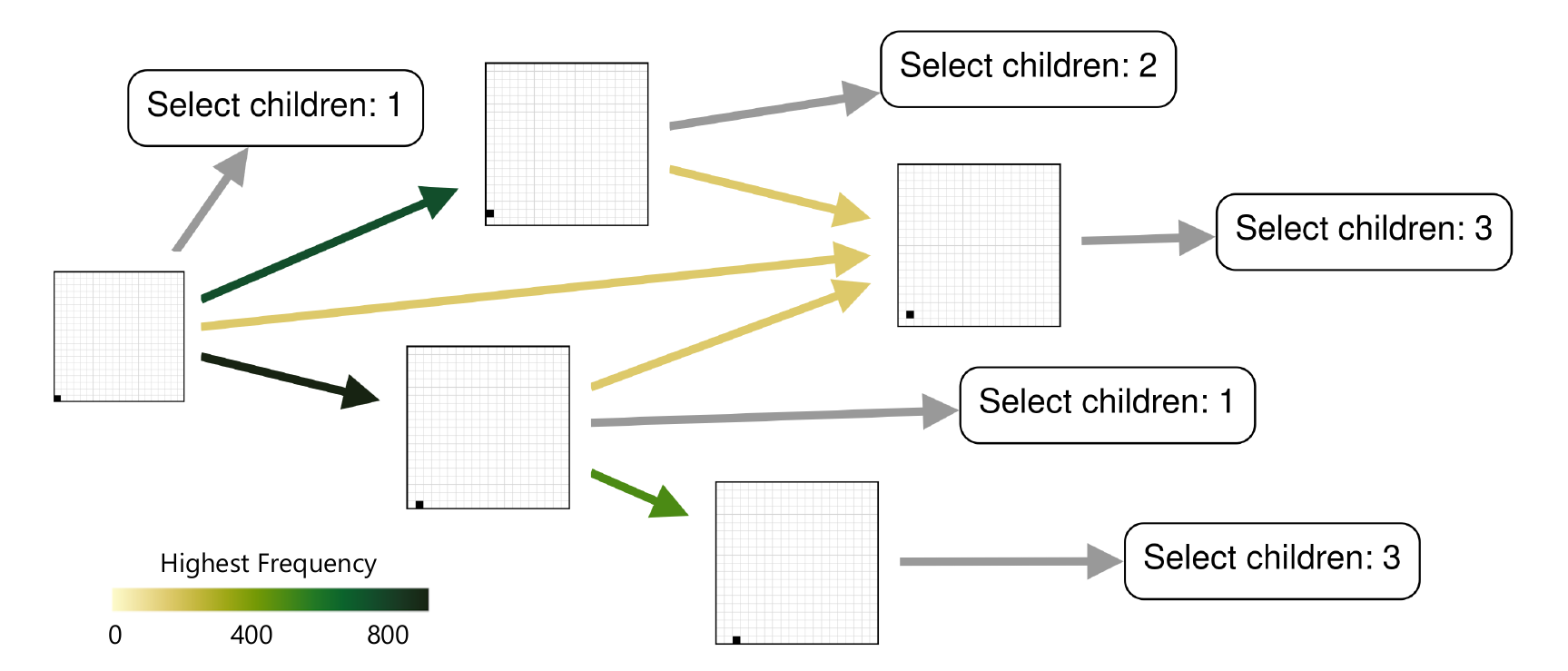}
  \caption{\emph{DAG View} for the grid example, displaying a subgraph of the DAG accumulated from individual training sample trajectories. Selecting the aggregated children node expands the graph and reveals trajectories of interest. The nodes show states as positions within the grid. Edges are colored based on a metric chosen (in this case, frequency).
  }
  \label{fig:DAG}
\end{figure}

The \emph{DAG View} (see \cref{fig:DAG}) shows how samples are generated from sequences of actions. 
Logged trajectories are combined to construct a DAG that represents the various ways how samples can be constructed.
Since visualizing the full graph is neither computationally feasible nor informative, the graph is truncated, and interactive exploration mechanisms are provided.
This enables a targeted inspection of trajectories sampled by the model (\ref{task:traj-steps}).

To make the large DAG easier to interpret, several preprocessing steps are applied to the trajectory data.
Many trajectories contain linear chains (i.e., sequences of nodes with exactly one parent and one child).
These chains are aggregated into a single transition.
Identical transitions (i.e., multiple edges from state $s$ to $s'$) are merged and visualized as a single transition, colored based on an aggregation metric.
We expand on the available metrics shared between the \emph{DAG View} and the \emph{Transition Heatmap} in Section~\ref{sec:transition_heatmap}.
Subgraphs can be explored in two ways: (\textit{i})~by interactively expanding the graph from the root, or (\textit{ii})~by selecting elements in other visualizations (see, for example, the interaction descriptions in \cref{sec:ranking,sec:projection}).

By default, only the root node (i.e., the source state) is displayed with an aggregated node indicating the number of children.
Selecting the \emph{children} node opens a table of child nodes and their attributes.
Users can then select children from the table to add them as dedicated nodes in the graph---each with their own \emph{children} nodes for further expansion.
The incremental construction of the graph allows users to investigate regions of interest in the sample space.
Selecting a sample node displays all complete trajectories passing through that sample's state, enabling rapid exploration of local subgraphs.
This exploration of trajectory formation helps users understand potential errors that, for example, prohibit or over-emphasize certain transitions (\ref{task:traj-factors}).
Selecting objects in other visualizations shows their corresponding trajectories in the \emph{DAG View}.

Most existing approaches on interactive node expansion focus on trees (see \cref{sec:graph_exploration}).
In trees, collapsing or expanding a node affects only one subtree.
DAGs, however, give rise to the multiple-parent problem:
since a node may have multiple parents and shared descendants, it is unclear which nodes should appear or disappear and how to handle layout stability upon interaction.
\emph{Expansion} proved to be straightforward: a node may be listed as a child of multiple expanded nodes; upon selection, all transitions from an existing node to the new node are drawn.
However, when \emph{collapsing} a node, simply removing its transitions is not sufficient.
In a tree, children can be collapsed recursively.
In a DAG, however, we must also check whether a node is connected to other existing nodes
To improve performance for large DAGs, we compute the DAG structure from the trajectory data in advance and store it in a database.
This accelerates repeated querying.
Upon expanding the graph, new nodes and edges are added to the layout, possibly introducing edge crossings unless the layout is reoptimized. We decided on redrawing the graph after expansions to keep the layout clean, with the tradeoff that users have to reorient themselves.
In future work, we aim to implement this redrawing with an animated transition to help users keep orientation.

\subsection{Transition Heatmap}
\label{sec:transition_heatmap}

\begin{figure}[t]
  \centering
  \includegraphics[width=\columnwidth, alt={Left: Transition Heatmap showing transition rank versus iterations. Each time a transition is sampled, it is colored based on the chosen metric. Right: Transition hover information showing the source and target node and a line plot of the transition probabilities over iterations.}]{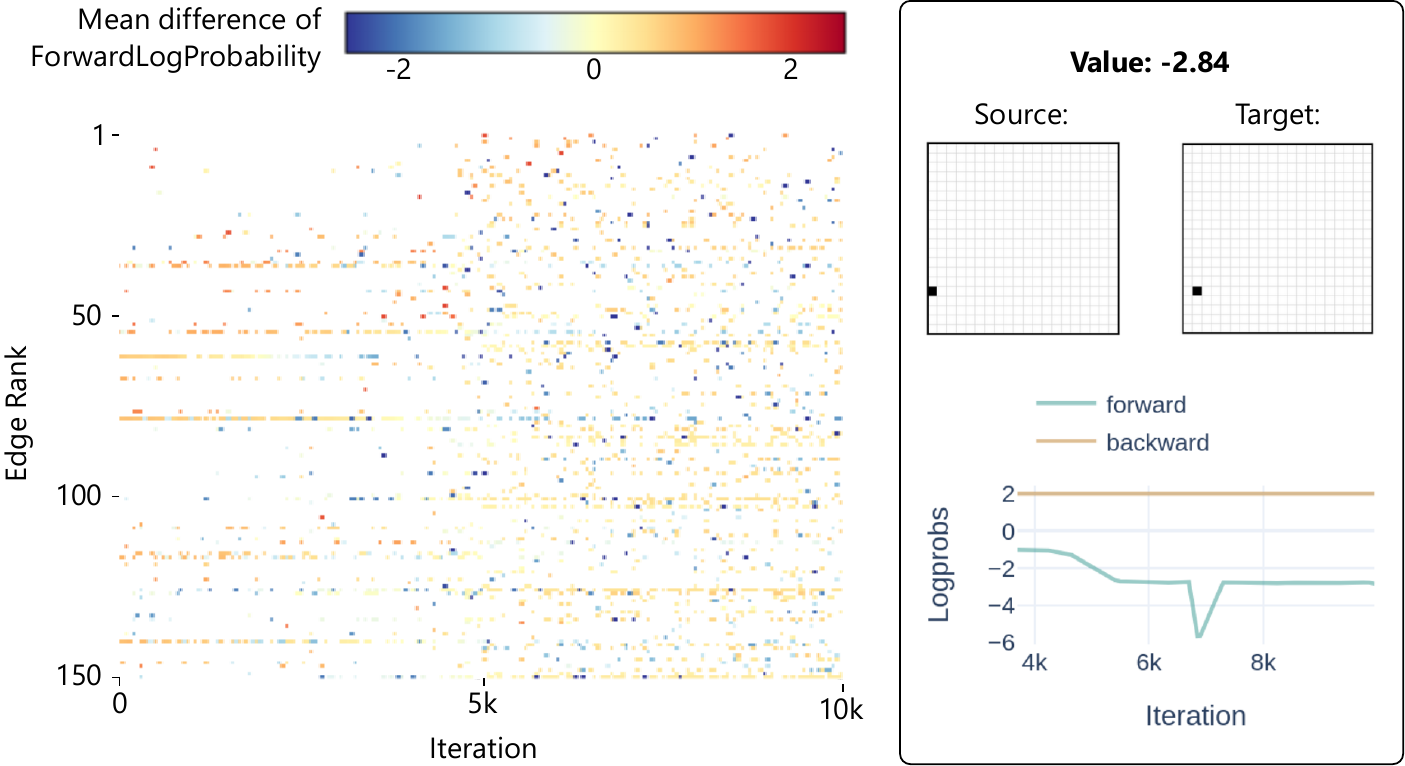}
  \subfigsCaption{\emph{Transition Heatmap} of the DAG (left). In this example, transitions are ranked and colored by the difference in transition probabilities. Transitions are only colored for the iterations they are sampled in. Hovering over a transition mark provides additional details~(right): customizable representations of source and target states, and a line chart showing the progression of the transition probabilities during training.}
  \label{fig:edge_heatmap}
\end{figure}

The \emph{Transition Heatmap} (see \cref{fig:edge_heatmap}) shows transition edges ranked by a specific metric and visualized across iterations.
The coloring by metric is consistent with the encoding used for the edges in the \emph{DAG View}.
Users can choose between the transition probability, the variance of the transition probabilities, or the transition frequency.
By default, the forward transitions are displayed, but users can also choose to inspect the backward transition probabilities.
The transition ranks are shown on the vertical axis, while the horizontal axis indicates the training iterations, giving an overview of the iterations in which each transition appears.
Hovering over a transition reveals details about its source and target states, as well as how transition probabilities evolved during training (\ref{task:dynamic-shifts}; see right part of \cref{fig:edge_heatmap}).
Selecting transitions in the \emph{Transition Heatmap} highlights all samples containing these transitions as part of their trajectories in the \emph{Sample Ranking} and the \emph{State Projection}.
The entire associated trajectories are shown in the \emph{DAG View}.
Analyzing transition probabilities allows validating model behavior with domain expert knowledge, especially in conjunction with the customizable domain-specific sample representations.

\subsection{Implementation and Environment Integration}
\label{sec:impl}

\begin{figure*}[t]
    \centering
    \includegraphics[width=\textwidth, alt={State projection, Transition Heatmap, and DAG View for the grid example use case.}]{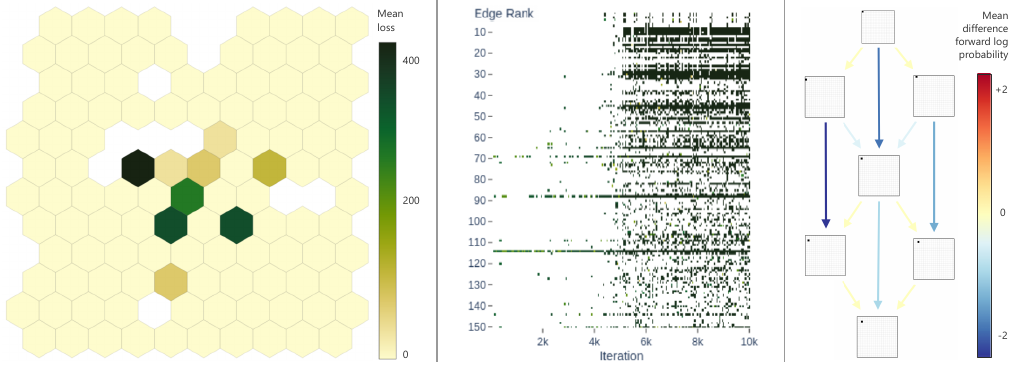}
    \caption{Select \emph{GFlowState} visualizations for the grid case study. Left:~\emph{State Projection} with bins colored by mean loss. High-loss regions are located at the center of the grid. Middle:~The \emph{Transition Heatmap} showing the highest transition probabilities. Only after iteration 4,500, transitions closer to the edges of the grid are sampled (as inferred from hover details). The three transitions consistently sampled during early training (Rank 69, 89, and 112) correspond to transitions from the source state towards the bottom-left mode, which is close to the start state. Right:~A subgraph of the environment DAG. Transitions with decreasing probability during training are shown in blue. This reveals the decreasing relevance of the first mode in the bottom-left corner of the grid as new high-reward modes are explored---a~sign that the GFlowNet learns to represent the reward better by sampling actions more diversely.}
    \label{fig:use-case-grid}
\end{figure*}

We implemented \emph{GFlowState} in Python using Plotly Dash\footnote{\url{https://dash.plotly.com/}}.
We chose a Python-based solution to enable easy integration into the developer workflow.
Training samples are stored in an SQL database, with two linked tables for edges and nodes.
Using a database allows \emph{GFlowState} to scale even to long training runs.

\emph{GFlowState} is designed for easy integration into custom environments.
Users only need to provide environment-specific implementations for four functions:
 \begin{enumerate}
     \item \emph{Database string format}---Users need to specify how to convert each state into a unique string to be saved in the database.
     \item \emph{Feature representation}---To allow an application of standard dimension reduction algorithms for the \emph{State Projection} view, users need to specify a feature representation for states (e.g., using the features provided in the GFlowNet policy).
     \item \emph{Single-state visual representation}---This visual representation is used to indicate states in the \emph{DAG View} and in the on-hover details of the other views. It can be either a figure or text.
     \item \emph{Multi-state visual representation}---This visual or textual representation of a set of states is used to summarize multiple states within selected hexbins in the detail view of the \emph{State Projection}.
 \end{enumerate}

The source code for \emph{GflowState} is openly available at GitHub~\cite{Holeczek_GFlowNet_Explorer}.
The repository contains a tutorial on how to use and interpret \emph{GFlowState}, and demo datasets for exploring it. Additional information is available in the repository's README file.

\section{Case Studies}

We evaluate \emph{GFlowState} through two case studies conducted in collaboration with researchers at Mila, who are actively developing GFlowNets.
One of the researchers (also a co-author in this paper) performed an in-depth analysis of both case studies.
In addition, we collected high-level feedback from three other researchers for the \emph{Crystals} environment.

\subsection{Grid Environment}

We recorded a training run of the grid environment (see \cref{sec:background}), where we trained a GFlowNet for 10,000 iterations on a $20\times20$ grid.
The reward function has four modes, one near each corner of the grid (see \cref{fig:reward}).

The \emph{Sample Ranking} proved helpful for localizing changes.
The expert started their analysis in the \emph{Sample Ranking} view.
Here, he immediately identified iterations in which new modes of the reward function were discovered (see \cref{fig:ranking}).
Mode discoveries stand out in the \emph{Sample Ranking} because many new objects are moving into the high-reward positions, and other object rankings drop: this behavior leads to a pattern of many lines moving down in the ranking visualization.
Initially, the model started sampling from the bottom-left mode---closest to the source state.
After around iteration 4,500, the model discovered the top-right mode, and after around iteration 5,000, the remaining two modes.
According to the expert, these sudden discoveries of high-reward areas often appear in GFlowNet training.
In practice, identifying these areas previously required repeatedly loading checkpoints and inspecting the generated samples.
\emph{GFlowState} accelerates this process.

The expert moved on to the \emph{State Projection}, describing it as valuable for assessing model quality.
In particular, he highlighted the correlation between rewards and forward transition probabilities as important. Using an additional validation set of all grid points, the expert could quickly identify which areas the model discovered at each training time step.
The areas with the highest loss appeared to be in the center of the grid (\cref{fig:use-case-grid}, left).
According to the expert, this was expected, as the model rarely samples from this low-reward area and therefore has difficulty modeling it accurately.

Looking at the \emph{Transition Heatmap}, the expert identified a common feature of the highest transition probabilities: most high-ranking transitions were sampled only from iteration 4,500 onward (\cref{fig:use-case-grid}, middle).
A closer inspection based on the on-hover detail view revealed that they correspond to transitions of states near the edges of the grid.
This confirmed the intuition that transitions towards the newly discovered high-reward modes (top left and bottom right in the grid) must have high flow to consistently reach these states. 

Finally, the expert used the \emph{DAG View} to identify the transitions with the highest variance over iterations.
He quickly found that transitions towards the bottom-left mode decreased in probability over time, accounting for newly discovered modes and the decreasing importance of the mode discovered first. (\cref{fig:use-case-grid}, right).
This further corroborated the notion that the GFlowNet moves away from only sampling from the high-reward mode closest to the starting point and instead learns to create a more diverse set of states that more accurately represent the true reward landscape.
The expert concluded that the \emph{DAG View} was helpful for analyzing trajectories, which had previously been a tedious process.
He appreciated the ability to explore subgraphs, which was described by the expert as intuitive and fast.


\subsection{Crystals Environment}

\begin{figure*}[t]
  \centering
  \includegraphics[width=.75\textwidth, alt={Sample Ranking based on highest reward for the Crystal use case. The highest ranking samples are discovered early on, parallel lines show no new high reward discoveries later on.}]{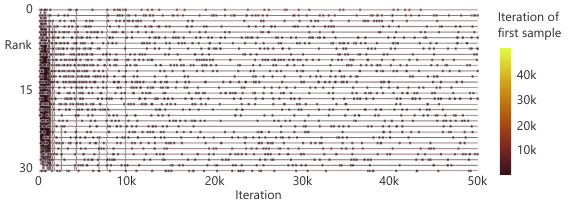}
  \caption{\emph{Sample Ranking} for the crystal environment showing the highest reward samples. The color indicates at what point in training the sample was first added to the ranking. The parallel horizontal lines over wide stretches of the training indicate that most of the high-ranked samples are discovered early.}
  \label{fig:crystal_ranking}
\end{figure*}

While the grid environment is a useful and intuitive example, we wanted to test \emph{GFlowState} in an application-oriented setting.
To this end, we used the \emph{crystal} environment as proposed by the Mila AI4Science team~\cite{mistal_crystal-gfn_2023}.
A~crystal structure can be described by three main properties: (\textit{i})~its composition, which defines the elements present and their stoichiometry, (\textit{ii})~its space group, which describes the crystal symmetry, and (\textit{iii})~its lattice parameters, which specify the lengths and angles of the unit cell.
A~GFlowNet can sequentially generate these properties.

For this case study, we restricted the composition of the crystal to platinum and palladium, which are widely used catalytic metals.
We considered only the cubic space groups 225 and 229 and limited the lattice parameter ranges to physically plausible ranges~\footnote{The full configuration can be found here: \url{https://github.com/alexhernandezgarcia/gflownet/blob/main/config/experiments/crystals/turaco_density.yaml}}.
This results in a constrained but meaningful search space, while still keeping training costs low.
As reward, we used crystal density, which is defined as the total mass of all atoms divided by the cell volume based on the lattice parameters.
This yields a simple property that can be derived computationally from the structure of the generated crystal.
We recorded a training run of 50,000 iterations.
For comparison, we used a validation dataset of crystals that spans the entire sample space of the environment, which was feasible due to the constraints.
As the lattice parameters were continuous, we discretized them to make the DAG meaningful.
We invited four researchers with backgrounds in machine learning, materials science, or both, to interpret the results in \emph{GFlowState}.
All of them actively work with GFlowNets for crystal generation.

\begin{figure*}[t]
    \centering
    \includegraphics[width=\textwidth, alt={Three versions of the State Projection for the crystals environment.}]{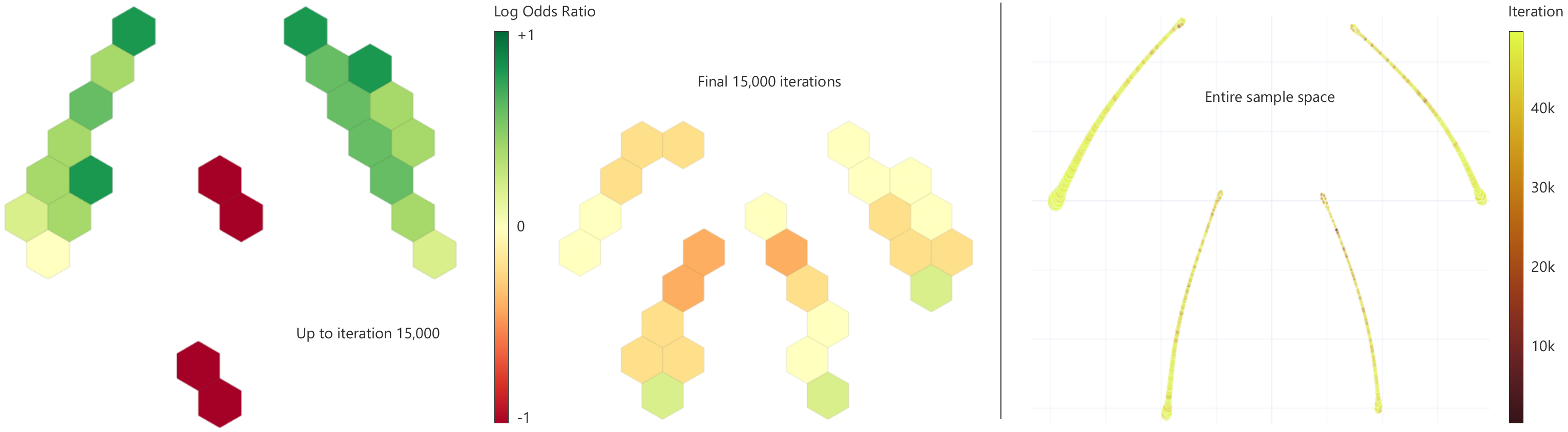}
    \caption{Three variations of the \emph{State Projection} view for the crystals environment. Left:~Aggregated view of the samples generated up to iteration 15,000. The red hexagons indicate areas in the validation set not yet covered by the samples. Middle:~Aggregated view of the samples generated  in the last 15,000 iterations. Coloring shows that the distributions of samples and validation set objects are now much more similar across the whole sample space. Right:~All generated samples as a scatterplot without aggregation. The four lines correspond to the four generated compositions. Lattice length decreases towards the bottom, resulting in higher crystal density and reward.}
    \label{fig:crystal-projection-v2}
\end{figure*}

\begin{figure*}[t]
  \centering
  \includegraphics[width=\textwidth, alt={DAG View showing a subgraph of the DAG for the Crystal environment}]{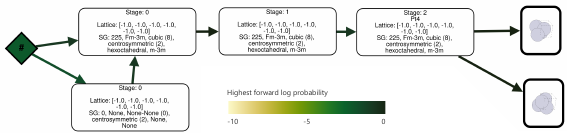}
  \caption{\emph{DAG View} showing a subgraph of the DAG for the crystal environment. The DAG was expanded to show the trajectories that lead to the two  structures with the highest reward.}
  \label{fig:crystal_DAG}
\end{figure*}

Looking at the \emph{Sample Ranking}, the experts quickly identified that one generated crystal structure (Platinum2 with space group 225 and lattice lengths close to 2) dominated the high-reward region~\cref{fig:crystal_ranking}.
The model discovered this structure early on.
However, by looking at the \emph{State Projection} for an early iteration (see~\cref{fig:crystal-projection-v2}, left), the experts noticed that at this point large parts of the sample space had remained unexplored (as evidenced by the odds ratios based on a comparison with the validation set).
The experts highlighted the importance of the \emph{Sample Ranking} as a fast way to inspect the training dynamics and identify iterations of interest.
When investigating a later state of the training, the comparison with the validation set indicates that the model generalizes over the whole sample space better (see \cref{fig:crystal-projection-v2}, middle).

The experts resumed to explore the sample space in more detail, using the non-aggregated \emph{State Projection} (see \cref{fig:crystal-projection-v2}, right).
The sample space shows clear clusters of the generated structures based on the composition.
The lattice length spreads samples within each cluster.
A~shorter lattice length with the same composition corresponds to higher density, making high-reward areas easily identifiable at one end of each cluster.

The experts reported that the ability to explore the DAG of sample trajectories was the most helpful component of the system and also the one that benefited most from domain-expert knowledge (see \cref{fig:crystal_DAG}).
They explained that during the development of the crystal environment, they frequently analyzed a large number of trajectories.
They also noted that the \emph{GFlowState} interface could substantially accelerate their workflow for this process, providing a fast way for GFlowNet experts to debug trajectory generation and for domain experts to check the plausibility of the generated states.

The researchers also emphasized the importance of analyzing dependencies between subenvironments: In the evaluated crystal environment configuration, the space group is generated first, subsequently constraining the generation of the composition.
Being able to inspect in detail how the selected space group influences the resulting composition is helpful when training models to generate physically meaningful crystals.
The experts also highlighted the usefulness of exploring different stages of the training process.
Access to historical trajectories helps in understanding why the model behaves as it does.


\section{Limitations and Future Work}

While our collaborators found \emph{GFlowState} to be a generally useful solution for their analyses of GFlowNet training behaviors, the case studies also surfaced some areas with potential for improval.
In future work, we intend to improve the following aspects of \emph{GFlowState}.

\textbf{Transitions.} 
The states in the crystal environment are desribed by heterogenous multidimensional information: space group, composition, and lattice parameters.
For such complex state spaces, understanding the transition between two states is less intuitive than in other environments.
Currently, the \emph{DAG View} relies solely on the state representation.
Displaying the selected \emph{action} on each transition of the DAG in addition to the states could make the state transitions easier to interpret.

\textbf{Clustering.} Although experts appreciated the existing measures for reducing the size of the DAG, which improve interpretability, additional clustering mechanisms could be beneficial.
In particular, user-guided clustering based on various state features could provide a quick overview of how different properties influence trajectories.
Similarly, clustering high-reward samples in the ranking could help group similar objects, shifting focus from solely the reward to the diversity of generated samples.

\textbf{Continuous Environments.} Initially, our system was designed for discrete environments.
As shown in the application for crystal structure generation, \emph{GFlowState} can also handle continuous environments if they can be meaningfully discretized.
As of now, decisions regarding this discretization have to be taken early on when starting the training run.
This makes later changes to the precision with which to discretize impossible. 

By logging data in their continuous representation, we would allow dynamic adjustments to the precision of the discretization after the training.
This way, users could adjust the precision depending on the task.
For overview purposes, for example, choosing a lower precision could automatically clusters similar states and simplify the DAG, whereas a higher precision could allow for detailed inspection of states.
While promising, this idea also gives rise to new challenges, such as repeated graph computation.

\textbf{Streaming During Training.} Inspecting the DAG during training is theoretically possible, but it requires repeated expensive computation, slowing down training.
The main reason for the complexity is the truncation of linear chains (i.e., removing nodes of degree two) to make the graph more accessible.
This process requires a recomputation of the entire DAG.
By providing users with the untruncated graph and an option to truncate it on demand, we hope to find a good balance between a low impact on training and full support of our system during training runtime.


\textbf{Correlation Estimation.} A key metric highlighted by all experts is the correlation between forward transition probabilities and reward, which they consider an important indicator of model performance.
In the current setup, this metric is computed from training samples, which are collected at different iterations, and visualized for subsets of the sample space. However, in large sample spaces, later training iterations may not provide sufficient samples to reliably estimate this correlation.

One possible solution is to compute the forward transition probabilities for objects in the validation dataset and use them to estimate the metric. This can be achieved by sampling backward trajectories for each dataset object using importance sampling, and then applying the trained model to obtain the corresponding forward transition probabilities.
This approach would allow estimating the correlation between reward and forward transition probabilities across the sample space for different iterations without relying on the training samples.
The main drawback is the additional complexity, as it requires either storing model checkpoints or performing additional computations during training.
This contrasts \emph{GFlowState}'s current lightweight approach, which depends only on already generated samples.
While adding complexity to the setup of our system, implementing this strategy as an option for users could greatly improve assessing model quality, especially as experts highlighted its value in the case studies.


\section{Conclusion}

We introduced \emph{GFlowState}, an interactive visual analytics system designed to analyze GFlowNet training.
The system exposes aspects of GFlowNet training that are difficult to observe with existing monitoring tools, including the structure of sample trajectories, computation of relevant metrics for various parts of the sample space, and overall training dynamics.
By combining interactive DAG exploration of trajectories, state-projection visualizations, and rankings of generated objects, the system enables developers to investigate exploration behavior, identify underexplored regions of the sample space, and better understand how transition probabilities evolve throughout training. 

Through case studies in both the intuitive grid environment and an application-oriented crystal structure generation task, we demonstrated how the system helps researchers interpret training dynamics, detect potential shortcomings in model behavior, and analyze trajectory generation in detail.
Feedback from experts highlighted the usefulness of interactive trajectory exploration and local analysis of the relationship between forward transition probabilities and reward, both of which are difficult to assess using existing tools.

More broadly, our work illustrates the value of model-specific visual analytics systems for machine learning paradigms.
As GFlowNets continue to gain traction in scientific discovery applications, systems that support their interpretability will become increasingly important.
By making the structural dynamics of GFlowNet training observable, \emph{GFlowState} provides developers with new opportunities to better understand, diagnose, and improve GFlowNets in practical applications.

\section*{Supplemental Materials}
\label{sec:supplemental_materials}
The code for \emph{GFlowNet} is openly available on GitHub\footnote{\url{https://github.com/florianholeczek/GFlowNet_Training_Vis_Pilot}}. 
Information about integration of custom environments can be found in the repository. The project is currently in the process of being integrated into the GFlowNet library\footnote{\url{https://github.com/alexhernandezgarcia/gflownet}}.
An online demo is available at \url{http://gflowstate.jku-vds-lab.at/}.

\acknowledgments{
	This research was supported by the Austrian Science Fund under grant number FWF DFH 23–N, the Austrian Research Promotion Agency (FFG 881844), and the ETH AI Center through an ETH AI Center postdoctoral fellowship to Christina Humer.
    A big thanks to our collaborators at Mila for their valuable feedback.
}

\bibliographystyle{abbrv-doi-hyperref}

\bibliography{main}

@String{procNeurIPS = {Advances in Neural Information Processing Systems}}

@String{jourTVCG = {{IEEE} Transactions on Visualization and Computer Graphics}}

@String{jourNature = {Nature}}

@String{jourACMComputSurv = {ACM Computing Surveys}}

@String{jourJMLR = {Journal of Machine Learning Research}}

@String{procPMLR = {Proceedings of the International Conference on Machine Learning}}

@String{jourCGF = {Computer Graphics Forum}}

@String{jourTIIS = {{ACM} Transactions on Interactive Intelligent Systems}}

@article{lee_treeplus_2006,
	title = {{TreePlus}: Interactive Exploration of Networks with Enhanced Tree Layouts},
	volume = {12},
	issn = {1077-2626},
	shorttitle = {{TreePlus}},
	url = {http://ieeexplore.ieee.org/document/1703363/},
	doi = {10.1109/TVCG.2006.106},
	number = {6},
	urldate = {2018-11-12},
	journal = jourTVCG,
	author = {Lee, Bongshin and Parr, C.S. and Bederson, B.B. and Veksler, V.D. and Gray, W.D. and Kotfila, C. and Kotfila, C.},
	year = {2006},
	pages = {1414--1426},
 }

@article{nobre2019juniper,
	title = {Juniper: A Tree+Table Approach to Multivariate Graph Visualization},
	volume = {25},
	xcopyright = {https://ieeexplore.ieee.org/Xplorehelp/downloads/license-information/IEEE.html},
	issn = {1077-2626, 1941-0506, 2160-9306},
	shorttitle = {Juniper},
	url = {https://ieeexplore.ieee.org/document/8454344/},
	doi = {10.1109/TVCG.2018.2865149},
	number = {1},
	urldate = {2018-04-11},
	journal = jourTVCG,
	author = {Nobre, Carolina and Streit, Marc and Lex, Alexander},
	month = jan,
	year = {2019},
	pages = {544 -- 554},
}

@article{wang2023scientific,
	title = {Scientific discovery in the age of artificial intelligence},
	volume = {620},
	copyright = {2023 Springer Nature Limited},
	issn = {1476-4687},
	url = {https://www.nature.com/articles/s41586-023-06221-2},
	doi = {10.1038/s41586-023-06221-2},
	number = {7972},
	urldate = {2024-09-12},
	journal = jourNature,
	xpublisher = {Nature Publishing Group},
	author = {Wang, Hanchen and Fu, Tianfan and Du, Yuanqi and Gao, Wenhao and Huang, Kexin and Liu, Ziming and Chandak, Payal and Liu, Shengchao and Van Katwyk, Peter and Deac, Andreea and Anandkumar, Anima and Bergen, Karianne and Gomes, Carla P. and Ho, Shirley and Kohli, Pushmeet and Lasenby, Joan and Leskovec, Jure and Liu, Tie-Yan and Manrai, Arjun and Marks, Debora and Ramsundar, Bharath and Song, Le and Sun, Jimeng and Tang, Jian and Veličković, Petar and Welling, Max and Zhang, Linfeng and Coley, Connor W. and Bengio, Yoshua and Zitnik, Marinka},
	xmonth = aug,
	year = {2023},
	pages = {47--60},
}

@article{jain2023gflownets,
	title = {{GFlowNets} for {AI}-driven scientific discovery},
	volume = {2},
	issn = {2635-098X},
	url = {https://pubs.rsc.org/en/content/articlelanding/2023/dd/d3dd00002h},
	doi = {10.1039/D3DD00002H},
	number = {3},
	urldate = {2026-03-13},
	journal = {Digital Discovery},
	publisher = {RSC},
	author = {Jain, Moksh and Deleu, Tristan and Hartford, Jason and Liu, Cheng-Hao and Hernandez-Garcia, Alex and Bengio, Yoshua},
	xmonth = jun,
	year = {2023},
	pages = {557--577},
}

@article{rolnick2022tackling,
	title = {Tackling Climate Change with Machine Learning},
	volume = {55},
	issn = {0360-0300},
	url = {https://dl.acm.org/doi/10.1145/3485128},
	doi = {10.1145/3485128},
	number = {2},
	urldate = {2024-10-21},
	journal = jourACMComputSurv,
	author = {Rolnick, David and Donti, Priya L. and Kaack, Lynn H. and Kochanski, Kelly and Lacoste, Alexandre and Sankaran, Kris and Ross, Andrew Slavin and Milojevic-Dupont, Nikola and Jaques, Natasha and Waldman-Brown, Anna and Luccioni, Alexandra Sasha and Maharaj, Tegan and Sherwin, Evan D. and Mukkavilli, S. Karthik and Kording, Konrad P. and Gomes, Carla P. and Ng, Andrew Y. and Hassabis, Demis and Platt, John C. and Creutzig, Felix and Chayes, Jennifer and Bengio, Yoshua},
	xmonth = feb,
	year = {2022},
	pages = {42:1--42:96},
}

@article{dara2022machine,
	title = {Machine Learning in Drug Discovery: A Review},
	volume = {55},
	issn = {1573-7462},
	shorttitle = {Machine Learning in Drug Discovery},
	url = {https://doi.org/10.1007/s10462-021-10058-4},
	doi = {10.1007/s10462-021-10058-4},
	number = {3},
	urldate = {2026-03-13},
	journal = {Artificial Intelligence Review},
	author = {Dara, Suresh and Dhamercherla, Swetha and Jadav, Surender Singh and Babu, CH Madhu and Ahsan, Mohamed Jawed},
	xmonth = mar,
	year = {2022},
	pages = {1947--1999},
}

@article{bengio2023gflownet,
	title = {{GFlowNet} Foundations},
	volume = {24},
	url = {http://jmlr.org/papers/v24/22-0364.html},
    note = {\url{http://jmlr.org/papers/v24/22-0364.html}},
	number = {210},
	journal = jourJMLR,
	author = {Bengio, Yoshua and Lahlou, Salem and Deleu, Tristan and Hu, Edward J. and Tiwari, Mo and Bengio, Emmanuel},
	year = {2023},
	pages = {1--55},
}

@inproceedings{bengio2021flow,
	address = {Red Hook, NY, USA},
	xseries = {{NIPS} '21},
	title = {Flow network based generative models for non-iterative diverse candidate generation},
	xisbn = {978-1-7138-4539-3},
	urldate = {2026-03-16},
    url = {https://proceedings.neurips.cc/paper_files/paper/2021/hash/e614f646836aaed9f89ce58e837e2310-Abstract.html},
    note = {\url{https://proceedings.neurips.cc/paper_files/paper/2021/hash/e614f646836aaed9f89ce58e837e2310-Abstract.html}},
	booktitle = procNeurIPS,
	xpublisher = {Curran Associates Inc.},
	author = {Bengio, Emmanuel and Jain, Moksh and Korablyov, Maksym and Precup, Doina and Bengio, Yoshua},
	xmonth = dec,
    volume = {34},
	year = {2021},
	pages = {27381--27394},
}

@inproceedings{malkin2022trajectory,
	title = {Trajectory balance: Improved credit assignment in {GFlowNets}},
	volume = {35},
	url = {https://proceedings.neurips.cc/paper_files/paper/2022/file/27b51baca8377a0cf109f6ecc15a0f70-Paper-Conference.pdf},
	note = {\url{https://proceedings.neurips.cc/paper_files/paper/2022/file/27b51baca8377a0cf109f6ecc15a0f70-Paper-Conference.pdf}},
	booktitle = procNeurIPS,
	xpublisher = {Curran Associates, Inc.},
	author = {Malkin, Nikolay and Jain, Moksh and Bengio, Emmanuel and Sun, Chen and Bengio, Yoshua},
	editor = {Koyejo, S. and Mohamed, S. and Agarwal, A. and Belgrave, D. and Cho, K. and Oh, A.},
	year = {2022},
	pages = {5955--5967},
}

@inproceedings{shen2023understanding,
	title = {Towards Understanding and Improving GFlowNet Training},
	issn = {2640-3498},
	note = {\url{https://proceedings.mlr.press/v202/shen23a.html}},
	urldate = {2026-03-16},
	booktitle = procPMLR,
	xpublisher = {PMLR},
	author = {Shen, Max W. and Bengio, Emmanuel and Hajiramezanali, Ehsan and Loukas, Andreas and Cho, Kyunghyun and Biancalani, Tommaso},
	xmonth = jul,
	year = {2023},
	pages = {30956--30975},
}

@inproceedings{yin_synergy_2025,
	xseries = {{AAAI}'25/{IAAI}'25/{EAAI}'25},
	title = {Synergy of {GFlowNet} and protein language model makes a diverse antibody designer},
	volume = {39},
	isbn = {978-1-57735-897-8},
	url = {https://doi.org/10.1609/aaai.v39i21.34370},
	doi = {10.1609/aaai.v39i21.34370},
	urldate = {2026-03-16},
	booktitle = {Proceedings of the Thirty-Ninth {AAAI} Conference on Artificial Intelligence},
	author = {Yin, Mingze and Zhou, Hanjing and Zhu, Yiheng and Wu, Jialu and Wu, Wei and Li, Mingyang and Fu, Kun and Wang, Zheng and Hsieh, Chang-Yu and Hou, Tingjun and Wu, Jian},
	xmonth = feb,
	year = {2025},
	pages = {22164--22172},
}

@inproceedings{podina_catalyst_2025,
	title = {Catalyst {GFlowNet} for electrocatalyst design: A hydrogen evolution reaction case study},
	shorttitle = {Catalyst {GFlowNet} for electrocatalyst design},
    booktitle={AI for Accelerated Materials Design},
	note = {\url{https://openreview.net/forum?id=z80sbzFMBE}},
	urldate = {2025-11-19},
	author = {Podina, Lena and Hernández-García, Alex and Humer, Christina and Duval, Alexandre and Schmidt, Victor and Ramlaoui, Ali and Chatterjee, Shahana and Bengio, Yoshua and Rolnick, David and Therrien, Félix},
	xmonth = oct,
	year = {2025},
}

@article{chanussot_open_2021,
	title = {Open Catalyst 2020 ({OC20}) Dataset and Community Challenges},
	volume = {11},
	url = {https://doi.org/10.1021/acscatal.0c04525},
	doi = {10.1021/acscatal.0c04525},
	number = {10},
	urldate = {2026-03-16},
	journal = {ACS Catalysis},
	xpublisher = {American Chemical Society},
	author = {Chanussot, Lowik and Das, Abhishek and Goyal, Siddharth and Lavril, Thibaut and Shuaibi, Muhammed and Riviere, Morgane and Tran, Kevin and Heras-Domingo, Javier and Ho, Caleb and Hu, Weihua and Palizhati, Aini and Sriram, Anuroop and Wood, Brandon and Yoon, Junwoong and Parikh, Devi and Zitnick, C. Lawrence and Ulissi, Zachary},
	xmonth = may,
	year = {2021},
	pages = {6059--6072},
}

@inproceedings{mistal_crystal-gfn_2023,
	title = {Crystal-{GFN}: sampling materials with desirable properties and constraints},
	shorttitle = {Crystal-{GFN}},
    booktitle = {AI for Accelerated Materials Design},
	note = {\url{https://openreview.net/forum?id=l167FjdPOv}},
	urldate = {2026-03-16},
	author = {Mistal and Hernández-García, Alex and Volokhova, Alexandra and Duval, Alexandre AGM and Bengio, Yoshua and Sharma, Divya and Carrier, Pierre Luc and Koziarski, Michał and Schmidt, Victor},
	xmonth = nov,
	year = {2023},
}

@inproceedings{zahavy_graying_2016,
	title = {Graying the black box: Understanding {DQNs}},
	issn = {1938-7228},
	shorttitle = {Graying the black box},
	note = {\url{https://proceedings.mlr.press/v48/zahavy16.html}},
	urldate = {2026-03-16},
	booktitle = procPMLR,
	xpublisher = {PMLR},
	author = {Zahavy, Tom and Ben-Zrihem, Nir and Mannor, Shie},
	xmonth = jun,
	year = {2016},
	pages = {1899--1908},
}

@article{metz_visitor_2023,
	title = {{VISITOR}: Visual Interactive State Sequence Exploration for Reinforcement Learning},
	volume = {42},
	issn = {1467-8659},
	shorttitle = {{VISITOR}},
	url = {https://onlinelibrary.wiley.com/doi/abs/10.1111/cgf.14839},
	doi = {10.1111/cgf.14839},
	number = {3},
	urldate = {2026-03-16},
	journal = jourCGF,
	author = {Metz, Yannick and Bykovets, Eugene and Joos, Lucas and Keim, Daniel and El-Assady, Mennatallah},
	year = {2023},
	pages = {397--408},
}

@article{strobelt_seq2seq-vis_2019,
	title = {Seq2seq-{Vis}: A Visual Debugging Tool for Sequence-to-Sequence Models},
	volume = {25},
	issn = {1941-0506},
	shorttitle = {Seq2seq-{Vis}},
	url = {https://ieeexplore.ieee.org/abstract/document/8494828},
	doi = {10.1109/TVCG.2018.2865044},
	number = {1},
	urldate = {2024-05-23},
	journal = jourTVCG,
	author = {Strobelt, Hendrik and Gehrmann, Sebastian and Behrisch, Michael and Perer, Adam and Pfister, Hanspeter and Rush, Alexander M.},
	xmonth = jan,
	year = {2019},
	pages = {353--363},
}

@article{strobelt_lstmvis_2018,
	title = {{LSTMVis}: A Tool for Visual Analysis of Hidden State Dynamics in Recurrent Neural Networks},
	volume = {24},
	issn = {1941-0506},
	shorttitle = {{LSTMVis}},
	url = {https://ieeexplore.ieee.org/abstract/document/8017583},
	doi = {10.1109/TVCG.2017.2744158},
	number = {1},
	urldate = {2024-05-23},
	journal = jourTVCG,
	author = {Strobelt, Hendrik and Gehrmann, Sebastian and Pfister, Hanspeter and Rush, Alexander M.},
	xmonth = jan,
	year = {2018},
	pages = {667--676},
}

@article{sevastjanova_visual_2023,
	title = {Visual Comparison of Language Model Adaptation},
	volume = {29},
	issn = {1941-0506},
	url = {https://ieeexplore.ieee.org/document/9904461},
	doi = {10.1109/TVCG.2022.3209458},
	number = {1},
	urldate = {2026-03-16},
	journal = jourTVCG,
	author = {Sevastjanova, Rita and Cakmak, Eren and Ravfogel, Shauli and Cotterell, Ryan and El-Assady, Mennatallah},
	xmonth = jan,
	year = {2023},
	pages = {1178--1188},
}

@article{wang_cnn_2021,
	title = {{CNN} {Explainer}: Learning Convolutional Neural Networks with Interactive Visualization},
	volume = {27},
	issn = {1941-0506},
	shorttitle = {{CNN} {Explainer}},
	url = {https://ieeexplore.ieee.org/abstract/document/9222325},
	doi = {10.1109/TVCG.2020.3030418},
	number = {2},
	urldate = {2024-05-23},
	journal = jourTVCG,
	author = {Wang, Zijie J. and Turko, Robert and Shaikh, Omar and Park, Haekyu and Das, Nilaksh and Hohman, Fred and Kahng, Minsuk and Polo Chau, Duen Horng},
	xmonth = feb,
	year = {2021},
	pages = {1396--1406},
}

@inproceedings{spinner_revealing_2025,
	address = {Vienna, Austria},
	title = {Revealing the Unwritten: Visual Investigation of Beam Search Trees to Address Language Model Prompting Challenges},
	isbn = {979-8-89176-253-4},
	shorttitle = {Revealing the Unwritten},
	url = {https://aclanthology.org/2025.acl-demo.29/},
	doi = {10.18653/v1/2025.acl-demo.29},
	urldate = {2026-03-16},
	booktitle = {Proceedings of the 63rd Annual Meeting of the Association for Computational Linguistics ({Volume} 3: System Demonstrations)},
	xpublisher = {Association for Computational Linguistics},
	author = {Spinner, Thilo and Sevastjanova, Rita and Kehlbeck, Rebecca and Stähle, Tobias and Keim, Daniel A. and Deussen, Oliver and Spitz, Andreas and El-Assady, Mennatallah},
	xeditor = {Mishra, Pushkar and Muresan, Smaranda and Yu, Tao},
	xmonth = jul,
	year = {2025},
	pages = {295--306},
}

@article{spinner_-generaitor_2024,
	title = {-{generAItor}: Tree-in-the-loop Text Generation for Language Model Explainability and Adaptation},
	volume = {14},
	issn = {2160-6455},
	shorttitle = {-{generAItor}},
	url = {https://dl.acm.org/doi/10.1145/3652028},
	doi = {10.1145/3652028},
	number = {2},
	urldate = {2026-03-16},
	journal = jourTIIS,
	author = {Spinner, Thilo and Kehlbeck, Rebecca and Sevastjanova, Rita and Stähle, Tobias and Keim, Daniel A. and Deussen, Oliver and El-Assady, Mennatallah},
	xmonth = jun,
	year = {2024},
	pages = {14:1--14:32},
}

@misc{Holeczek_GFlowNet_Explorer,
    author = {Holeczek, Florian},
    license = {GPL-3.0},
    title = {{GFlowState}},
    note = {\url{https://github.com/florianholeczek/GFlowNet_Training_Vis_Pilot}}
}

@article{hohman_visual_2019,
    title = {Visual Analytics in Deep Learning: An Interrogative Survey for the Next Frontiers},
    volume = {25},
    issn = {1941-0506},
    shorttitle = {Visual Analytics in Deep Learning},
    doi = {10.1109/TVCG.2018.2843369},
    number = {8},
    journal = jourTVCG,
    author = {Hohman, Fred and Kahng, Minsuk and Pienta, Robert and Chau, Duen Horng},
    xmonth = aug,
    year = {2019},
    pages = {2674--2693},
}

@article{holeczek_gflownet_2025,
    title = {{GFlowNet} {Playground} - Theory and Examples for an Intuitive Understanding},
    note = {\url{https://gfn-playground.jku-vds-lab.at/}},
    journal = {Workshop on Visualization for AI Explainability},
    author = {Holeczek, Florian and Hillisch, Alexander and Hinterreiter, Andreas and Hernandez-Garcia, Alex and Streit, Marc and Humer, Christina},
    year = {2025},
}

@inproceedings{puhringer_instanceflow_2020,
    address = {Salt Lake City},
    title = {{InstanceFlow}: Visualizing the Evolution of Classifier Confusion at the Instance Level},
    doi = {10.1109/VIS47514.2020.00065},
    booktitle = {2020 {IEEE} Visualization Conference ({VIS})},
    xpublisher = {IEEE},
    author = {Pühringer, Michael and Hinterreiter, Andreas and Streit, Marc},
    year = {2020},
    pages = {291--295},
}

@article{hinterreiter_confusionflow_2022,
    title = {{ConfusionFlow}: A Model-Agnostic Visualization for Temporal Analysis of Classifier Confusion},
    volume = {28},
    issn = {1941-0506},
    shorttitle = {{ConfusionFlow}},
    doi = {10.1109/TVCG.2020.3012063},
    number = {2},
    journal = jourTVCG,
    author = {Hinterreiter, Andreas and Ruch, Peter and Stitz, Holger and Ennemoser, Martin and Bernard, Jürgen and Strobelt, Hendrik and Streit, Marc},
    xmonth = feb,
    year = {2022},
    xnote = {Conference Name: IEEE Transactions on Visualization and Computer Graphics},
    pages = {1222--1236},
}

@article{kahng_gan_2019,
    title = {{GAN} {Lab}: Understanding Complex Deep Generative Models using Interactive Visual Experimentation},
    volume = {25},
    issn = {1941-0506},
    shorttitle = {{GAN} {Lab}},
    url = {https://ieeexplore.ieee.org/document/8440049},
    doi = {10.1109/TVCG.2018.2864500},
    number = {1},
    urldate = {2026-03-19},
    journal = jourTVCG,
    author = {Kahng, Minsuk and Thorat, Nikhil and Chau, Duen Horng and Viégas, Fernanda B. and Wattenberg, Martin},
    xmonth = jan,
    year = {2019},
    pages = {310--320},
}

@article{prasad_evolved_2025,
    title = {{EvolvED}: Evolutionary Embeddings to Understand the Generation Process of Diffusion Models},
    xvolume = {n/a},
    xcopyright = {© 2025 The Author(s). Computer Graphics Forum published by Eurographics - The European Association for Computer Graphics and John Wiley \& Sons Ltd.},
    issn = {1467-8659},
    shorttitle = {{EvolvED}},
    url = {https://onlinelibrary.wiley.com/doi/abs/10.1111/cgf.70301},
    doi = {10.1111/cgf.70301},
    xlanguage = {en},
    xnumber = {n/a},
    urldate = {2026-03-19},
    journal = jourCGF,
    author = {Prasad, Vidya and van Gorp, Hans and Humer, Christina and van Sloun, Ruud J. G. and Vilanova, Anna and Pezzotti, Nicola},
    xmonth = dec,
    year = {2025},
    xnote = {\_eprint: https://onlinelibrary.wiley.com/doi/pdf/10.1111/cgf.70301},
    xpages = {e70301},
}

@article{wang_visual_2022,
    title = {Visual Analytics for {RNN}-Based Deep Reinforcement Learning},
    volume = {28},
    issn = {1941-0506},
    url = {https://ieeexplore.ieee.org/document/9420254/},
    doi = {10.1109/TVCG.2021.3076749},
    number = {12},
    urldate = {2026-03-19},
    journal = jourTVCG,
    author = {Wang, Junpeng and Zhang, Wei and Yang, Hao and Yeh, Chin-Chia Michael and Wang, Liang},
    xmonth = dec,
    year = {2022},
    pages = {4141--4155},
}

@inproceedings{mishra_why_2022,
    title = {Why? Why not? When? Visual Explanations of Agent Behaviour in Reinforcement Learning},
    issn = {2165-8773},
    shorttitle = {Why?},
    url = {https://ieeexplore.ieee.org/document/9787897/},
    doi = {10.1109/PacificVis53943.2022.00020},
    urldate = {2026-03-19},
    booktitle = {2022 {IEEE} 15th Pacific Visualization Symposium ({PacificVis})},
    author = {Mishra, Aditi and Soni, Utkarsh and Huang, Jinbin and Bryan, Chris},
    xmonth = apr,
    year = {2022},
    note = {ISSN: 2165-8773},
    pages = {111--120},
}

@misc{tensorflow_developers_tensorflow_2026,
    title = {{TensorFlow}},
    xcopyright = {Apache License 2.0},
    url = {https://zenodo.org/doi/10.5281/zenodo.4724125},
    doi = {10.5281/ZENODO.4724125},
    urldate = {2026-03-20},
    publisher = {Zenodo},
    author = {TensorFlow Developers},
    xmonth = mar,
    year = {2026},
}

@misc{biewald_experiment_2020,
    title = {Experiment Tracking with Weights and Biases},
    url = {https://www.wandb.com/},
    author = {Biewald, Lukas},
    year = {2020},
}

@article{von_landesberger_visual_2011,
    title = {Visual Analysis of Large Graphs: State-of-the-Art and Future Research Challenges},
    volume = {30},
    xcopyright = {© 2011 The Authors Computer Graphics Forum © 2011 The Eurographics Association and Blackwell Publishing Ltd.},
    issn = {1467-8659},
    shorttitle = {Visual {Analysis} of {Large} {Graphs}},
    url = {https://onlinelibrary.wiley.com/doi/abs/10.1111/j.1467-8659.2011.01898.x},
    doi = {10.1111/j.1467-8659.2011.01898.x},
    xlanguage = {en},
    number = {6},
    urldate = {2026-03-27},
    journal = jourCGF,
    author = {von Landesberger, T. and Kuijper, A. and Schreck, T. and Kohlhammer, J. and van Wijk, J.j. and Fekete, J.-D. and Fellner, D.w.},
    year = {2011},
    xnote = {\_eprint: https://onlinelibrary.wiley.com/doi/pdf/10.1111/j.1467-8659.2011.01898.x},
    pages = {1719--1749},
}

@article{yuan2021survey,
  title={A survey of visual analytics techniques for machine learning},
  author={Yuan, Jun and Chen, Changjian and Yang, Weikai and Liu, Mengchen and Xia, Jiazhi and Liu, Shixia},
  journal={Computational Visual Media},
  volume={7},
  number={1},
  pages={3--36},
  year={2021},
  doi={10.1007/s41095-020-0191-7}
}

@article{eckelt_visual_2022,
    title = {Visual Exploration of Relationships and Structure in Low-Dimensional Embeddings},
    issn = {1941-0506},
    doi = {10.1109/TVCG.2022.3156760},
    journal = {IEEE Transactions on Visualization and Computer Graphics},
    author = {Eckelt, Klaus and Hinterreiter, Andreas and Adelberger, Patrick and Walchshofer, Conny and Dhanoa, Vaishali and Humer, Christina and Heckmann, Moritz and Steinparz, Christian and Streit, Marc},
    year = {2022},
    xpages = {1--1},
}

\appendix 
\crefalias{section}{appendix} 

\end{document}